\documentclass[runningheads]{llncs}
\usepackage{graphicx}
\usepackage{amsmath,amssymb} 
\usepackage{color}
\usepackage[width=122mm,left=12mm,paperwidth=146mm,height=193mm,top=12mm,paperheight=217mm]{geometry}

\usepackage{algorithm}
\usepackage{algpseudocode}
\algrenewcommand\algorithmicindent{.9em}%

\usepackage{bbm}
\usepackage{comment}
\usepackage[british,UKenglish,USenglish,english,american]{babel}
\usepackage{multirow}

\newcommand{\eg}{\rm{e.g.}}

\usepackage[pagebackref=true,breaklinks=true,letterpaper=true,colorlinks,bookmarks=false]{hyperref}

\begin{document}
\pagestyle{headings}
\mainmatter

\title{Towards High Performance \\ Video Object Detection for Mobiles} 

\titlerunning{Towards High Performance Video Object Detection for Mobiles}

\authorrunning{Towards High Performance Video Object Detection for Mobiles}

\author{Xizhou Zhu\thanks{This work is done when Xizhou Zhu and Xingchi Zhu are interns at Microsoft Research Asia.} \quad Jifeng Dai \quad Xingchi Zhu$^\star$ \quad Yichen Wei \quad Lu Yuan}
\institute{Microsoft Research Asia \\ \hspace{-0.1in}{\tt\small \{v-xizzhu,jifdai,v-xizh14,luyuan,yichenw\}@microsoft.com}}

\maketitle

\vspace{-0.2em}
\begin{abstract}
	Despite the recent success of video object detection on Desktop GPUs, its architecture is still far too heavy for mobiles. It is also unclear whether the key principles of sparse feature propagation and multi-frame feature aggregation apply at very limited computational resources. In this paper, we present a light weight network architecture for video object detection on mobiles. Light weight image object detector is applied on sparse key frames. A very small network, Light Flow, is designed for establishing correspondence across frames. A flow-guided GRU module is designed to effectively aggregate features on key frames. For non-key frames, sparse feature propagation is performed. The whole network can be trained end-to-end. The proposed system achieves 60.2\% mAP score on ImageNet VID validation at speed of 25.6 fps on mobiles (e.g., HuaWei Mate 8).
\end{abstract}

\section{Introduction}

Object detection has achieved significant progress in recent years using deep neural networks~\cite{huang2016speed}. The general trend has been to make deeper and more complicated object detection networks~\cite{girshick2014rich,he2014spatial,girshick2015fast,ren2015faster,dai2016rfcn,lin2016fpn,he2017mask,dai2017deformable,liu2016ssd,redmon2016yolo9000} in order to achieve higher accuracy. However, these advances in improving accuracy are not necessarily making networks more efficient with respect to size and speed. In many real world applications, such as robotics, self-driving car, augmented reality, and mobile phone, the object detection tasks need to be carried out in a real-time fashion on a computationally limited platform.

Recently, there has been rising interest in building very small, low latency models that can be easily matched to the design requirements for mobile and embedded vision application, for example, SqueezeNet~\cite{iandola2016squeezenet}, MobileNet~\cite{howard2017mobilenets}, and ShuffleNet~\cite{zhang2017shufflenet}. These structures are general, but not specifically designed for object detection tasks. For this purpose, several small deep neural network architectures for object detection in static images are explored, such as YOLO~\cite{redmon2016you}, YOLOv2~\cite{redmon2016yolo9000}, Tiny YOLO~\cite{darknet13}, Tiny SSD~\cite{wong2018tiny}. However, directly applying these detectors to videos faces new challenges. First, applying the deep networks on all video frames introduces unaffordable computational cost. Second, recognition accuracy suffers from deteriorated appearances in videos that are seldom observed in still images, such as motion blur, video defocus, rare poses, etc.

\begin{figure}
	\begin{center}
		\includegraphics[width=.8\linewidth]{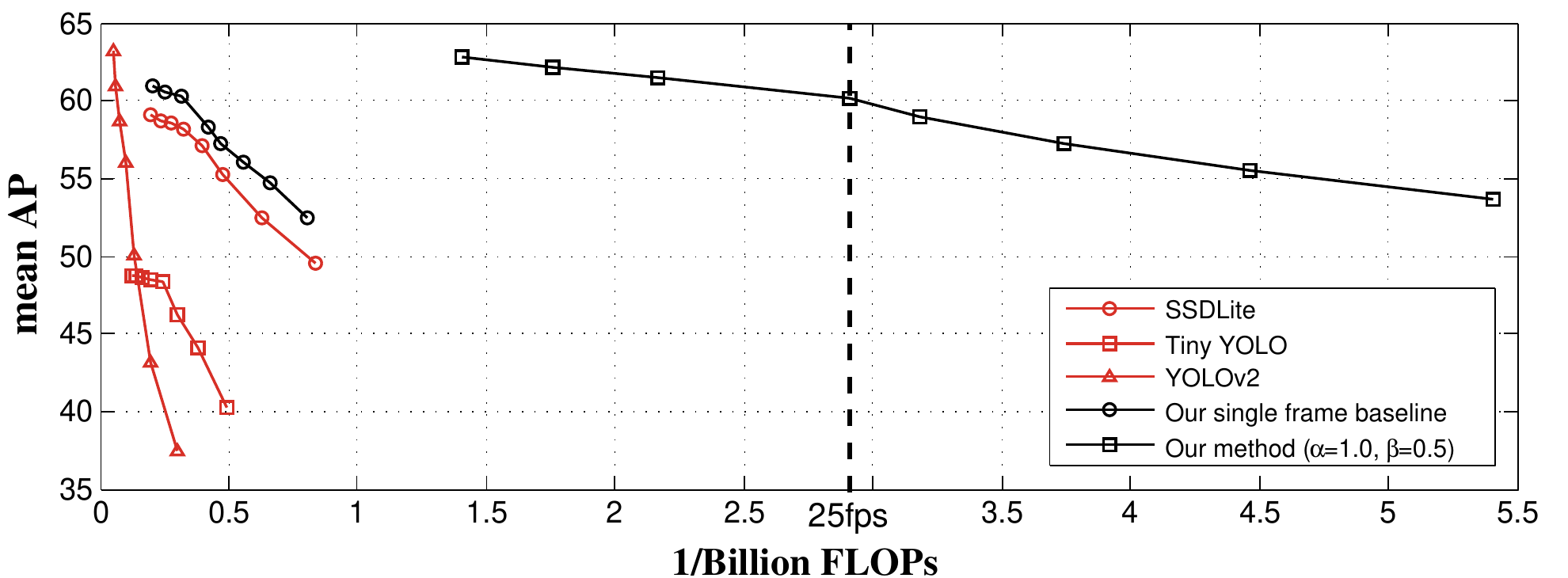}
	\end{center}
	\vspace{-1.5em}
	\caption{Speed-accuracy trade-off for different lightweight object detectors. Curves are drawn with varying image resolutions. Inference time is evaluated with TensorFlow Lite~\cite{tensorflow2015-whitepaper} on a single 2.3GHz Cortex-A72 processor of Huawei Mate 8.}
	\vspace{-1.5em}
	\label{fig.single_image_detectors}
\end{figure}

To address these issues, the current best practice~\cite{zhu2016dff,zhu2017flow,zhu2017towards} exploits temporal information for speedup and improvement on detection accuracy for videos. On one hand, \emph{sparse feature propagation} is used in~\cite{zhu2016dff,zhu2017towards} to save expensive feature computation on most frames. Features on these frames are propagated from sparse key frame cheaply. On the other hand, \emph{multi-frame feature aggregation} is performed in~\cite{zhu2017flow,zhu2017towards} to improve feature quality and detection accuracy.

Built on the two principles, the latest work~\cite{zhu2017towards} provides a good speed-accuracy tradeoff on Desktop GPUs. Unfortunately, the architecture is not friendly for mobiles. For example, flow estimation, as the key and common component in feature propagation and aggregation~\cite{zhu2016dff,zhu2017flow,zhu2017towards}, is still far from the demand of real-time computation on mobiles. Aggregation with long-term dependency is also restricted by limited runtime memory of mobiles.

This paper describes a light weight network architecture for mobile video object detection. It is primarily built on the two principles -- propagating features on majority non-key frames while computing and aggregating features on sparse key frames. However, we need to carefully redesign both structures for mobiles by considering speed, size and accuracy. On all frames, we present Light Flow, a very small deep neural network to estimate feature flow, which offers instant availability on mobiles. On sparse key frame, we present \emph{flow-guided Gated Recurrent Unit (GRU) based feature aggregation}, an effective aggregation on a memory-limited platform. Additionally, we also exploit a light image object detector for computing features on key frame, which leverage advanced and efficient techniques, such as depthwise separable convolution~\cite{sifre2014rigid} and Light-Head R-CNN~\cite{li2017light}.

The proposed techniques are unified to an end-to-end learning system. Comprehensive experiments show that the model steadily pushes forward the performance (speed-accuracy trade-off) envelope, towards high performance video object detection on mobiles. For example, we achieve $60.2\%$ mAP score on ImageNet VID validation at speed of 25.6 frame per second on mobiles (\eg, HuaWei Mate 8). It is one order faster than the best previous effort on fast object detection, with on par accuracy (see Figure~\ref{fig.single_image_detectors}). To the best of our knowledge, for the first time, we achieve realtime video object detection on mobile with reasonably good accuracy.

\section{Revisiting Video Object Detection Baseline}

Object detection in static images has achieved significant progress in recent years using deep CNN~\cite{huang2016speed}. State-of-the-art detectors share the similar network architecture, consisting of two conceptual steps. First step is feature network, which extracts a set of convolutional feature maps $F$ over the input image $I$ via a fully convolutional backbone network~\cite{simonyan2015very,szegedy2015going,he2016deep,szegedy2016inception,xie2017resnext,huang2016densely,chollet2016xception,howard2017mobilenets,zhang2017shufflenet}, denoted as ${\cal{N}}_{feat}(I) = F$. Second step is detection network, which generates detection result $y$ upon the feature maps $F$, by performing region classification and bounding box regression over either sparse object proposals~\cite{girshick2014rich,he2014spatial,girshick2015fast,ren2015faster,dai2016rfcn,lin2016fpn,he2017mask,dai2017deformable} or dense sliding windows~\cite{liu2016ssd,redmon2016you,redmon2016yolo9000,lin2017focal}, via a multi-branched sub-network, namely ${\cal{N}}_{det}(F) = y$. It is randomly initialized and jointly trained with ${\cal{N}}_{feat}$.

Directly applying these detectors to video object detection faces challenges from two aspects. For speed, applying single image detectors on all video frames is not efficient, since the backbone network ${\cal{N}}_{feat}$ is usually deep and slow. For accuracy, detection accuracy suffers from deteriorated appearances in videos that are seldom observed in still images, such as motion blur, video defocus, rare poses.

Current best practice~\cite{zhu2016dff,zhu2017flow,zhu2017towards} exploits temporal information via \emph{sparse feature propagation} and \emph{multi-frame feature aggregation}  to address the speed and accuracy issues, respectively.

\subsubsection{Sparse Feature Propagation}

Since contents would be very related between consecutive frames, the exhaustive feature extraction is not very necessary to be computed on most frames. Deep feature flow~\cite{zhu2016dff} provides an efficient way, which computes expensive feature network at only sparse key frames (\eg, every $10^{th}$) and propagates key frame feature maps to majority non-key frames, which results $5\times$ speedup with minor drop in accuracy.

During inference, feature maps on any non-key frame $i$ are propagated from its preceding key frame $k$ by,
\begin{equation}
F_{k \rightarrow i} = \mathcal{W}(F_k, M_{i \rightarrow k}),
\label{eq.feature_propagation_dff}
\end{equation}
where $F_k = \mathcal{N}_{feat}(I_k)$ is the feature of key frame $k$, and $\mathcal{W}$ represents the differentiable bilinear warping function. The two dimensional motion field $M_{i \rightarrow k}$ between two frames $I_i$ and $I_k$ is estimated through a flow network $\mathcal{N}_{flow}(I_k, I_i) = M_{i\rightarrow k}$, which is much cheaper than ${\cal{N}}_{feat}$. 

\subsubsection{Multi-frame Feature Aggregation}

To improve detection accuracy, flow-guided feature aggregation (FGFA)~\cite{zhu2017flow} aggregates feature maps from nearby frames, which are aligned well through the estimated flow.

The aggregated feature maps $\hat{F}_i$ at frame $i$ is obtained as a weighted average of nearby frames feature maps,
\begin{equation}
\hat{F}_i=\sum_{k \in [i-r, i+r]} W_{k\rightarrow i}\odot F_{k\rightarrow i},
\label{eq.feature_aggregation_fgfa}
\end{equation}
where $\odot$ denotes element-wise multiplication, and the weight $W_{k\rightarrow i}$ is adaptively computed as the similarity	between the propagated feature maps $F_{k\rightarrow i}$ and the feature map $F_i$ at frame $i$.

To avoid dense aggregation on all frames, \cite{zhu2017towards} suggested \emph{sparsely recursive feature aggregation}, which operates only on sparse key frames. Such a way retain the feature quality from aggregation but reduce the computational cost as well. Specifically, given two succeeding key frames $k$ and $k'$, the aggregated feature at frame $k'$ is computed by,
\begin{equation}
\hat{F}_{k'} = W_{k\rightarrow k'} \odot \hat{F}_{k \rightarrow k'} + W_{k'\rightarrow k'} \odot F_{k'}.
\label{eq.sparse_recursive_aggregation}
\end{equation}

\subsection{Practice for Mobiles}

As the two principles, \emph{sparse feature propagation} and \emph{multi-frame feature aggregation}, yield the best practice towards high performance (speed and accuracy trade-off) video object detection~\cite{zhu2017towards} on Desktop GPUs. Instead, there are very limited computational capability and runtime memory on mobiles. Therefore, what are principles for mobiles should be explored. 

\begin{itemize}
	\item Feature extraction and aggregation only operate on sparse key frames; while lightweight feature propagation is performed on majority non-key frames.
	\item Flow estimation is the key to feature propagation and aggregation. However, flow networks ${\cal{N}}_{flow}$ used in~\cite{zhu2016dff,zhu2017flow,zhu2017towards} are still far from the demand of real-time processing on mobiles. Specifically, FlowNet~\cite{dosovitskiy2015flownet} is $11.8\times$ FLOPs of MobileNet~\cite{howard2017mobilenets} under the same input resolutions. Even the smallest FlowNet Inception used in~\cite{zhu2016dff} is $1.6\times$ more FLOPs. A more cheaper ${\cal{N}}_{flow}$ is so necessary.
	\item Feature aggregation should be operated on aligned feature maps according to flow. Otherwise, displacements caused by large object motion would cause severe errors to aggregation. Long-term dependency in aggregation is also favoured because more temporal information can be fused together for better feature quality.
	\item The backbone network of single image detector should be as small as possible, since we need it to compute features on sparse key frame.
\end{itemize}

\section{Model Architecture for Mobiles}

\begin{figure}
	\begin{center}
		\includegraphics[width=.8\linewidth]{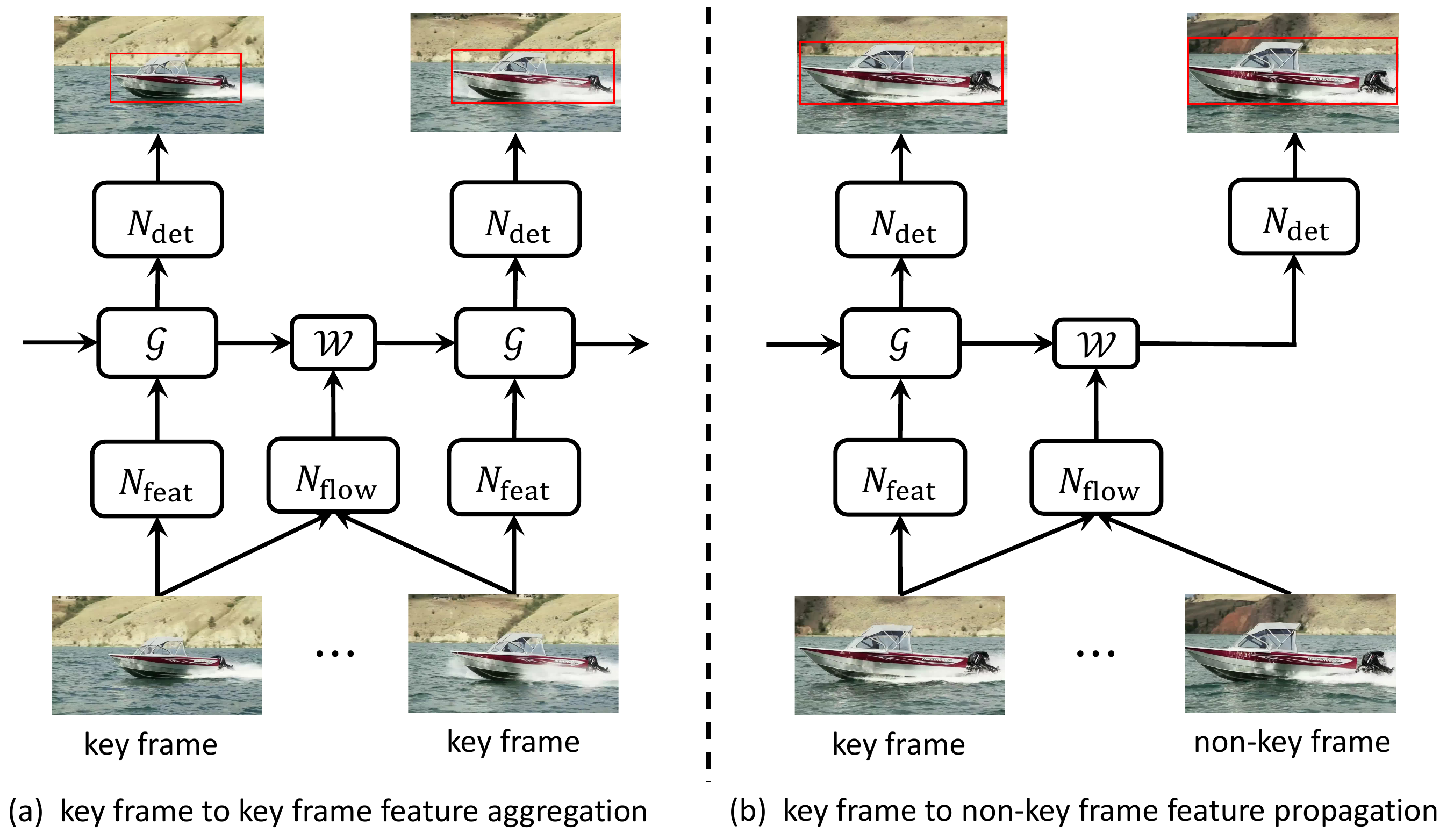}
	\end{center}
	\vspace{-1.5em}
	\caption{Illustration of video object detection for mobile by the proposed method. }
	\vspace{-1.5em}
	\label{fig.network}
\end{figure}

Based on the above principles, we design a much smaller network architecture for mobile video object detection. Inference pipeline is illustrated in Figure~\ref{fig.network}.

Given a key frame $k'$ and its proceeding key frame $k$, feature maps are first extracted by $F_{k'} = \mathcal{N}_{feat}(I_{k'})$, and then aggregated with its proceeding key frame aggregated feature maps $\hat{F}_k$ by,
\begin{equation}
\hat{F}_{k'} = \mathcal{G}(F_{k'}, \hat{F}_k,  M_{k' \rightarrow k}),
\label{eq.feature_aggregation}
\end{equation}
where $\mathcal{G}$ is a flow-guided feature aggregation function. The \emph{detection network} $\mathcal{N}_{det}$ is applied on $\hat{F}_{k'}$ to get detection predictions for the key frame $k'$.

Given a non-key frame $i$, the feature propagation from key frame $k$ to frame $i$ is denoted as,
\begin{equation}
\hat{F}_{k \rightarrow i} = \mathcal{W}(\hat{F}_{k}, M_{i \rightarrow k}),
\label{eq.feature_propagation}
\end{equation}
where $\hat{F}_{k}$ is the aggregated feature maps of key frame $k$, and $\mathcal{W}$ represents the differentiable bilinear warping function also used in~\cite{zhu2016dff}.
And then the \emph{detection network} $\mathcal{N}_{det}$ is applied on $\hat{F}_{k \rightarrow i}$ to get detection predictions for the non-key frame $i$.

Next, we will describe two new techniques which are specially designed for mobiles, including Light Flow, a more efficient flow network for mobiles, and a \emph{flow-guided GRU based feature aggregation} for better modeling long-term dependency, yielding better quality and accuracy.

\subsection{Light Flow}
\label{sec.lightflow}

FlowNet~\cite{dosovitskiy2015flownet} is originally proposed for pixel-level optical flow estimation. It is designed in a encoder-decoder mode followed by multi-resolution optical flow predictors. Two input RGB frames are concatenated to form a 6-channels input. In encoder, the input is converted into a bundle of feature maps in spatial dimensions to $1/64$ of input size through a class of convolutional layers. In decoder, the feature maps are fed to multiple deconvolution layers to achieve the high resolution flow prediction. After each deconvolution layer, the feature maps are concatenated with the last feature maps in encoder, which share the same spatial resolution and an upsampled coarse flow prediction. Multiple optical flow predictors follow each concatenated feature maps in decoder. Loss functions are applied to each predictor, but only the finest prediction is used during inference.

To speedup \emph{flow network} $\mathcal{N}_{flow}$ greatly, we present Light Flow, a more light weight flow network with several deliberate designs based on FlowNet~\cite{dosovitskiy2015flownet}. It only causes minor drop in accuracy ($15\%$ increasing in end-point error) but significantly speeds up by nearly $65\times$ theoretically (see Table.~\ref{table.ablation_flownet_epe_map}).

In encoder part, convolution is always the bottleneck of computation. Motivated by MobileNet~\cite{howard2017mobilenets}, we replace all convolutions to 3$\times$3 depthwise separable convolutions~\cite{sifre2014rigid} (each 3$\times$3 depthwise convolution followed by a 1$\times$1 pointwise convolution). Compared with standard 3$\times$3 convolution, the computation cost of 3$\times$3 depthwise separable convolution is reduced by $8\sim 9$ times with a slight drop in accuracy~\cite{howard2017mobilenets}.

In decoder part, each deconvolution operation is replaced by a nearest-neighbor upsampling followed by a depthwise separable convolution. \cite{odena2016deconvolution} replaces deconvolution with nearest-neighbor upsampling followed by a standard convolution to address checkerboard artifacts caused by deconvolution. By contrast, We leverage this idea, and further replace the standard convolution with depthwise separable convolution, to reduce computation cost.

Finally, we adopt a simple and effective way to consider multi-resolution predictions. It is inspired by FCN~\cite{evan2016fcn} which fuses multi-resolution semantic segmentation prediction as the final prediction in a explicit summation way. Unlike~\cite{dosovitskiy2015flownet}, we do not use only the finest optical flow prediction as final prediction during inference. Instead, multi-resolution predictions are up-sampled to the same spatial resolution with the finest prediction, and then are averaged as the final prediction. Also, during training, only a single loss function is applied on the averaged optical flow prediction instead of multiple loss functions after each prediction. Such a way can reduce end-point error by nearly $10\%$.

\subsubsection{Architecture and Implementation}

Network of Light Flow is illustrated in Table.~\ref{table.flownet_depthwise_architecture}. Each convolution operation is followed by batch normalization~\cite{ioffe2015batch} and Leaky ReLU nonlinearity~\cite{maas2013rectifier} with slope fixed as 0.1. Following~\cite{dosovitskiy2015flownet,ilg2016flownet2}, Light Flow is pre-trained on the Flying Chairs dataset. For training Light Flow, Adam~\cite{kingma2014adam} with a weight decay of 0.00004 is used as optimization method. 70k iterations are performed on 4 GPUs, with each GPU holding 64 image pairs.
A warm-up learning rate scheme is used in which we first train with a learning rate of 0.001 for 10k iterations. Then we train with learning rate of 0.01 for 20k iterations and divided it by 2 every 10k iterations.  

\setlength{\tabcolsep}{2pt}
\renewcommand{\arraystretch}{0.95}
\begin{table}[]
	\centering
	\small
	\begin{tabular}{l | l | c | l | l }
		\hline
		Name & Filter Shape & Stride & Output size & Input \\
		\hline
		\hline
		\multicolumn{5}{c}{Encoder} \\
		\hline
		\hline
		Images & & & 512$\times$384$\times$6& \\
		\hline
		Conv1\_dw & 3 $\times$ 3 $\times$ 6 dw & 2 & 256$\times$192$\times$6 & Images\\
		\hline
		Conv1 & 1 $\times$ 1 $\times$ 6 $\times$ 32 & 1 & 256$\times$192$\times$32 & Conv1\_dw\\
		\hline
		Conv2\_dw & 3 $\times$ 3 $\times$ 32 dw & 2 & 128$\times$96$\times$32 & Conv1\\
		\hline
		Conv2 & 1 $\times$ 1 $\times$ 32 $\times$ 64 & 1 & 128$\times$96$\times$64 & Conv2\_dw\\
		\hline
		Conv3\_dw & 3 $\times$ 3 $\times$ 64 dw & 2 & 64$\times$48$\times$64 & Conv2\\
		\hline
		Conv3 & 1 $\times$ 1 $\times$ 64 $\times$ 128 & 1 & 64$\times$48$\times$128 & Conv3\_dw\\
		\hline
		Conv4a\_dw & 3 $\times$ 3 $\times$ 128 dw & 2 & 32$\times$24$\times$128 & Conv3\\
		\hline
		Conv4a & 1 $\times$ 1 $\times$ 128 $\times$ 256 & 1 & 32$\times$24$\times$256 & Conv4a\_dw\\
		\hline
		Conv4b\_dw & 3 $\times$ 3 $\times$ 256 dw & 1 & 32$\times$24$\times$256 & Conv4a\\
		\hline
		Conv4b & 1 $\times$ 1 $\times$ 256 $\times$ 256 & 1 & 32$\times$24$\times$256 & Conv4b\_dw\\
		\hline
		Conv5a\_dw & 3 $\times$ 3 $\times$ 256 dw & 2 & 16$\times$12$\times$256 & Conv4b\\
		\hline
		Conv5a & 1 $\times$ 1 $\times$ 256 $\times$ 512 & 1 & 16$\times$12$\times$512 & Conv5a\_dw\\
		\hline
		Conv5b\_dw & 3 $\times$ 3 $\times$ 512 dw & 1 & 16$\times$12$\times$512 & Conv5a\\
		\hline
		Conv5b & 1 $\times$ 1 $\times$ 512 $\times$ 512 & 1 & 16$\times$12$\times$512 & Conv5b\_dw\\
		\hline
		Conv6a\_dw & 3 $\times$ 3 $\times$ 512 dw & 2 & 8$\times$6$\times$512 & Conv5b\\
		\hline
		Conv6a & 1 $\times$ 1 $\times$ 512 $\times$ 1024 & 1 & 8$\times$6$\times$1024 & Conv6a\_dw\\
		\hline
		Conv6b\_dw & 3 $\times$ 3 $\times$ 1024 dw & 1 & 8$\times$6$\times$1024 & Conv6a\\
		\hline
		Conv6b & 1 $\times$ 1 $\times$ 1024 $\times$ 1024 & 1 & 8$\times$6$\times$1024 & Conv6b\_dw\\
		\hline
		\hline
		\multicolumn{5}{c}{Decoder} \\
		\hline
		\hline
		Conv7\_dw & 3 $\times$ 3 $\times$ 1024 dw & 1 & 8$\times$6$\times$1024 & Conv6b\\
		\hline
		Conv7 & 1 $\times$ 1 $\times$ 1024 $\times$ 256 & 1 & 8$\times$6$\times$256 & Conv7\_dw\\
		\hline
		Conv8\_dw & 3 $\times$ 3 $\times$ 768 dw & 1 & 16$\times$12$\times$768 & [Conv7$\uparrow$, Conv5b]\\
		\hline
		Conv8 & 1 $\times$ 1 $\times$ 768 $\times$ 128 & 1 & 16$\times$12$\times$128 & Conv8\_dw\\
		\hline
		Conv9\_dw & 3 $\times$ 3 $\times$ 384 dw & 1 & 32$\times$24$\times$384 & [Conv8$\uparrow$, Conv4b]\\
		\hline
		Conv9 & 1 $\times$ 1 $\times$ 384 $\times$ 64 & 1 & 32$\times$24$\times$64 & Conv9\_dw\\
		\hline
		Conv10\_dw & 3 $\times$ 3 $\times$ 192 dw & 1 & 64$\times$48$\times$192 & [Conv9$\uparrow$, Conv3]\\
		\hline
		Conv10 & 1 $\times$ 1 $\times$ 192 $\times$ 32 & 1 & 64$\times$48$\times$32 & Conv10\_dw\\
		\hline
		Conv11\_dw & 3 $\times$ 3 $\times$ 96 dw & 1 & 128$\times$96$\times$96 & [Conv10$\uparrow$, Conv2]\\
		\hline
		Conv11 & 1 $\times$ 1 $\times$ 96 $\times$ 16 & 1 & 128$\times$96$\times$16 & Conv11\_dw\\
		\hline
		\hline
		\multicolumn{5}{c}{Optical Flow Predictors} \\
		\hline
		\hline
		Conv12\_dw & 3 $\times$ 3 $\times$ 256 dw & 1 & 8$\times$6$\times$256 & Conv7\\
		\hline
		Conv12 & 1 $\times$ 1 $\times$ 256 $\times$ 2 & 1 & 8$\times$6$\times$2 & Conv12\_dw\\
		\hline
		Conv13\_dw & 3 $\times$ 3 $\times$ 128 dw & 1 & 16$\times$12$\times$128 & Conv8\\
		\hline
		Conv13 & 1 $\times$ 1 $\times$ 128 $\times$ 2 & 1 & 16$\times$12$\times$2 & Conv13\_dw\\
		\hline
		Conv14\_dw & 3 $\times$ 3 $\times$ 64 dw & 1 & 32$\times$24$\times$64 & Conv9\\
		\hline
		Conv14 & 1 $\times$ 1 $\times$ 64 $\times$ 2 & 1 & 32$\times$24$\times$2 & Conv14\_dw\\
		\hline
		Conv15\_dw & 3 $\times$ 3 $\times$ 32 dw & 1 & 64$\times$48$\times$32 & Conv10\\
		\hline
		Conv15 & 1 $\times$ 1 $\times$ 32 $\times$ 2 & 1 & 64$\times$48$\times$2 & Conv15\_dw\\
		\hline
		Conv16\_dw & 3 $\times$ 3 $\times$ 16 dw & 1 & 128$\times$96$\times$16 & Conv11\\
		\hline
		Conv16 & 1 $\times$ 1 $\times$ 16 $\times$ 2 & 1 & 128$\times$96$\times$2 & Conv16\_dw\\
		\hline
		\hline
		\multicolumn{5}{c}{Multiple Optical Flow Predictions Fusion} \\
		\hline
		\hline
		\multirow{2}{*}{Average} & & & \multirow{2}{*}{128$\times$96$\times$2} & Conv12$\uparrow$$\uparrow$$\uparrow$$\uparrow$ + Conv13$\uparrow$$\uparrow$$\uparrow$ +\\
		& & & & Conv14$\uparrow$$\uparrow$ + Conv15$\uparrow$ + Conv16\\
		\hline
	\end{tabular}
	\vspace{0.1em}
	\caption{The details of the Light Flow architecture, 'dw' in filter shape denotes a depthwise separable convolution, $\uparrow$ is a 2$\times$ nearest neighbor upsampling, [$\cdot$,$\cdot$] is the concatenation operation.}
	\label{table.flownet_depthwise_architecture}
\end{table}

When applying Light Flow for our method, to get further speedup, two modifications are made.	First, following~\cite{zhu2016dff,zhu2017flow,zhu2017towards}, Light Flow is applied on images with half input resolution of the feature network, and has an output stride of 4. As the feature network has an output stride of 16, the flow field is downsampled to match the resolution of the feature maps. Second, since Light Flow is very small and has comparable computation with the detection network $\mathcal{N}_{det}$, sparse feature propagation is applied on the intermediate feature maps of the detection network (see Section~\ref{sec.detail_detector}, the 256-d feature maps in RPN~\cite{ren2015faster}, and the 490-d feature maps in Light-Head R-CNN~\cite{li2017light}), to further reduce computations for non-key frame.

\subsection{Flow-guided GRU based Feature Aggregation}

Previous works~\cite{zhu2017flow,zhu2017towards} have showed that feature aggregation plays an important role on improving detection accuracy. It should be explored how to learn complex and long-term temporal dynamics for a wide variety of sequence learning and prediction tasks. However, \cite{zhu2017flow} aggregates feature maps from nearby frame in a linear and memoryless way. Obviously, it only models short-term dependencies. Though recursive aggregation~\cite{zhu2017towards} has proven successful on fusing more past frames, it can be difficult to train it to learn long-term dynamics, likely due in part to the vanishing and exploding gradients problem that can result from propagating the gradients down through the many layers of the recurrent network.

Recently, \cite{chung2014empirical} has showed that Gated Recurrent Unit (GRU)~\cite{cho2014learning} is more powerful in modeling long-term dependencies than LSTM~\cite{hochreiter1997long} and RNN~\cite{elman1990finding}, because nonlinearities are incorporated into the network state updates. Inspired by this work, we incorporate convolutional GRU proposed by~\cite{ballas2015delving} into flow-guided feature aggregation function $\mathcal{G}$ instead of simply weighted average used in~\cite{zhu2017flow,zhu2017towards}. The aggregation function $\mathcal{G}$ in Eq.~\eqref{eq.feature_aggregation} is computed by,

\begin{equation}
\begin{split}
\hat{F}_{k'} &= \mathcal{G}(F_{k'}, \hat{F}_k,  M_{k' \rightarrow k}) \\
&= (1-z_t)\odot \hat{F}_{k\rightarrow k'} + z_t \odot \phi (W_h \star F_{k'} + U_h \star (r_t \odot \hat{F}_{k\rightarrow k'}) + b_h), \\			
z_t &= \sigma (W_z \star F_{k'} + U_z \star \hat{F}_{k\rightarrow k'} + b_z) \text{ is the update gate}, \\
r_t &= \sigma (W_r \star F_{k'} + U_r \star \hat{F}_{k\rightarrow k'} + b_r) \text{ is the reset gate}, \\	\end{split}
\label{eq.gru_aggregate}
\end{equation}
where $\hat{F}_{k \rightarrow k'} = \mathcal{W}(\hat{F}_{k}, M_{k' \rightarrow k})$, $W_\cdot, U_\cdot, b_\cdot$ are parameter tensors and vectors, $\odot$ denotes elementwise multiplication, $\star$ denotes $3\times3$ convolution, $\sigma$ is sigmoid function and $\phi$ is ReLU function.

Compared with the original GRU~\cite{cho2014learning}, there are three key differences.	First, $3\times3$ convolution is used instead of fully connected matrix multiplication, since fully connected matrix multiplication is too costly when GRU is applied to image feature maps. Second, $\phi$ is ReLU function instead of hyperbolic tangent function (tanh) for faster and better convergence. Third, we apply GRU only on sparse key frames (\eg, every $10^{th}$) instead of consecutive frames. Since two successive inputs for GRU would be apart 10 frames (1/3 second for a video with 30 fps), object displacements should be taken into account, and thus features should be aligned for GRU aggregation. By contrast, previous works~\cite{liu2017mobile,xingjian2015convolutional,li2016videolstm,ballas2015delving} based on either convolutional LSTM or convolutional GRU do not consider such a designing since they operate on consecutive frames instead, where object displacement would be small and neglected.

\subsection{Lightweight Key-frame Object Detector}\label{sec.detail_detector}

For key frame, we need a lightweight single image object detector, which consists of a feature network and a detection network. For the feature network, we adopt the state-of-the-art lightweight MobileNet~\cite{howard2017mobilenets} as the backbone network, which is designed for mobile recognition tasks. The MobileNet module is pre-trained on ImageNet classification task~\cite{russakovsky2015imagenet}. For the detection network, RPN~\cite{ren2015faster} and the recently presented Light-Head R-CNN~\cite{li2017light} are adopted, because of their light weight. Detailed implementation is illustrated below.

\subsubsection{Feature Network}
We remove the ending average pooling and the fully-connected layer of MobileNet~\cite{howard2017mobilenets}, and retain the convolutional layers. Since our input image resolution is very small (\eg, 224$\times$400), we increase feature resolution to get higher performance. First, a 3$\times$3 convolution is applied on top to reduce the feature dimension to 128, and then a nearest-neighbor upsampling is utilized to increase feature stride from 32 to 16. To give more detailed information, a 1$\times$1 convolution with 128 filters is applied to the last feature maps with feature stride 16, and then added to the upsampled 128-d feature maps.

\subsubsection{Detection Network}
RPN~\cite{ren2015faster} and Light-Head R-CNN~\cite{li2017light} are applied on the shared 128-d feature maps. In our model, to reduce computation of RPN, 256-d intermediate feature maps was utilized, which is half of originally used in~\cite{ren2015faster}. Three aspect ratios \{1:2, 1:1, 2:1\} and four scales \{$32^2$, $64^2$, $128^2$, $256^2$\} for RPN are set to cover objects with different shapes. For Light-Head R-CNN, a 1$\times$1 convolution with 10$\times$7$\times$7 filters was applied followed by a 7$\times$7 groups position-sensitive RoI warping~\cite{dai2016rfcn}. Then, two sibling fully connected layers are applied on the warped feature to predict RoI classification and regression.

\subsection{End-to-end Training}\label{sec.end2end_train}

All the modules in the entire architecture, including $\mathcal{N}_{feat}$, $\mathcal{N}_{det}$ and $\mathcal{N}_{flow}$, can be jointly trained for video object detection task.
In SGD, $n+1$ nearby video frames, {$I_i$, $I_k$, $I_{k-l}$, $I_{k-2l}$, ..., $I_{k-(n-1)l}$}, $0 \leq i - k < l$, are randomly sampled, where key frame duration $l = 10$ and key frame samples $n = 8$ are set for our experiments.
In the forward pass, $I_{k-(n-1)l}$ is assumed as a key frame, and the inference pipeline is exactly performed. Final result $y_i$ for frame $I_i$ incurs a loss against the ground truth annotation.
All operations are differentiable and thus can be end-to-end trained.

\section{Experiments}

Experiments are performed on ImageNet VID~\cite{russakovsky2015imagenet}, a large-scale benchmark for video object detection. Following the practice in~\cite{kang2016tcnn,lee2016multi}, model training and evaluation are performed on the 3,862 training video snippets and the 555 validation video snippets, respectively. The snippets are at frame rates of 25 or 30 fps in general. 30 object categories are involved, which are a subset of ImageNet DET annotated categories.

In training, following~\cite{kang2016tcnn,lee2016multi}, both the ImageNet VID training set and the ImageNet DET training set are utilized. In each mini-batch of SGD, either $n+1$ nearby video frames from ImageNet VID, or a single image from ImageNet DET, are sampled at 1:1 ratio. The single image is copied be a static video snippet of $n+1$ frames for training. In SGD, 240k iterations are performed on 4 GPUs, with each GPU holding one mini-batch. The learning rates are $10^{-3}$, $10^{-4}$ and $10^{-5}$ in the first 120k, the middle 60k and the last 60k iterations, respectively.

By default, the key-frame object detector is MobileNet+Light-Head R-CNN, and flow is estimated by Light Flow. The key frame duration length is every 10 frames. In both training and inference, the images are resized to a shorter side of 224 pixels and 112 pixels, for the image recognition network and  the flow network, respectively. Inference time is evaluated with TensorFlow Lite~\cite{tensorflow2015-whitepaper} on a single 2.3GHz Cortex-A72 processor of Huawei Mate 8. Theoretical computation is counted in FLOPs (floating point operations, note that a multiply-add is counted as 2 operations).

Following the practice in MobileNet~\cite{howard2017mobilenets}, two width multipliers, $\alpha$ and $\beta$, are introduced for controlling the computational complexity, by adjusting the network width. For each layer (except the final prediction layers) in $\mathcal{N}_{feat}$, $\mathcal{N}_{det}$ and $\mathcal{N}_{flow}$, its output channel number is multiplied by $\alpha$, $\alpha$ and $\beta$, respectively. The resulting network parameter number and theoretical computation change quadratically with the width multiplier. We experiment with $\alpha \in \{1.0, 0.75, 0.5\}$ and $\beta \in \{1.0, 0.75, 0.5\}$. By default, $\alpha$ and $\beta$ are set as 1.0.

\subsection{Ablation Study}

\subsubsection{Ablation on flow networks}

The middle panel of Table~\ref{table.ablation_flownet_epe_map} compares the proposed Light Flow with existing flow estimation networks on the Flying Chairs test set (384 x 512 input resolution). Following the protocol in~\cite{dosovitskiy2015flownet}, the accuracy is evaluated by the average end-point error (EPE). Compared with the original FlowNet design in~\cite{dosovitskiy2015flownet}, Light Flow ($\beta = 1.0$) can achieve $65.2\times$ theoretical speedup with $14.9\times$ less parameters. The flow estimation accuracy drop is small ($15\%$ relative increase in EPE). It is worth noting that it achieves higher accuracy than FlowNet Half and FlowNet Inception utilized in~\cite{zhu2016dff}, with at least one order less computation overhead. 
The speed of Light Flow can be further fastened with reduced network width, at certain cost of flow estimation accuracy. Flow estimation would not be a bottleneck in our mobile video object detection system.

Would such a light-weight flow network effectively guide feature propagation? To answer this question, we experiment with integrating different flow networks into our mobile video object detection system. The key-frame object detector is MobileNet+Light-Head R-CNN.

The rightmost panel of Table~\ref{table.ablation_flownet_epe_map} presents the results. The detection system utilizing Light Flow achieves accuracy very close to that utilizing the heavy-weight FlowNet (61.2\% v.s. 61.5\%), and is one order faster. Actually, the original FlowNet is so heavy that the detection system with FlowNet is even $2.7\times$ slower than simply applying the MobileNet+Light-Head R-CNN detector on each frame.

\setlength{\tabcolsep}{2pt}
\renewcommand{\arraystretch}{1}
\begin{table}[t]
	\begin{center}
		\begin{tabular}{c | c | c | c | c | c | c}
			\hline
			\multirow{2}{*}{flow network} & \multicolumn{3}{c|}{Flying Chairs \textit{test}} & \multicolumn{3}{c}{ImageNet VID \textit{validation}} \\
			\cline{2-7}
			& EPE & params (M) & FLOPs (B) &  mAP & params (M) & FLOPs (B)\\
			\hline
			FlowNet~\cite{dosovitskiy2015flownet}& 2.71 & 38.7 & 53.48 & 61.5 & 45.1 & 6.41\\
			\hline
			FlowNet Half~\cite{zhu2016dff}& 3.53 & 9.7 & 14.50 & - & - & - \\
			\hline
			FlowNet Inception~\cite{zhu2016dff}& 3.68 & 3.5 & 7.28 & - & - & - \\
			\hline
			FlowNet 2.0~\cite{ilg2016flownet2}& 1.71 & 162.5 & 269.39 & - & - & - \\
			\hline
			\hline
			1.0 Light Flow & 3.14 & 2.6 & 0.82 & 61.2 & 9.0 & 0.41\\
			\hline
			0.75 Light Flow & 3.63 & 1.4 & 0.48 & 60.6 & 7.8 & 0.37 \\
			\hline
			0.5 Light Flow & 4.44 & 0.7 & 0.23 & 60.1 & 7.1 & 0.34 \\
			\hline
		\end{tabular}
	\end{center}
	\caption{Ablation of different flow networks for optical flow prediction on Flying Chairs and for video object detection on ImageNet VID.}
	\vspace{-1.5em}
	\label{table.ablation_flownet_epe_map}
\end{table}

\subsubsection{Ablation on feature aggregation}

How important is to exploit flow to align features across frames? To answer this question, we experiment with a degenerated version of our method, where no flow-guided feature propagation is applied before aggregating features across key frames.

\begin{figure}
	\begin{center}
		\includegraphics[width=0.8\linewidth]{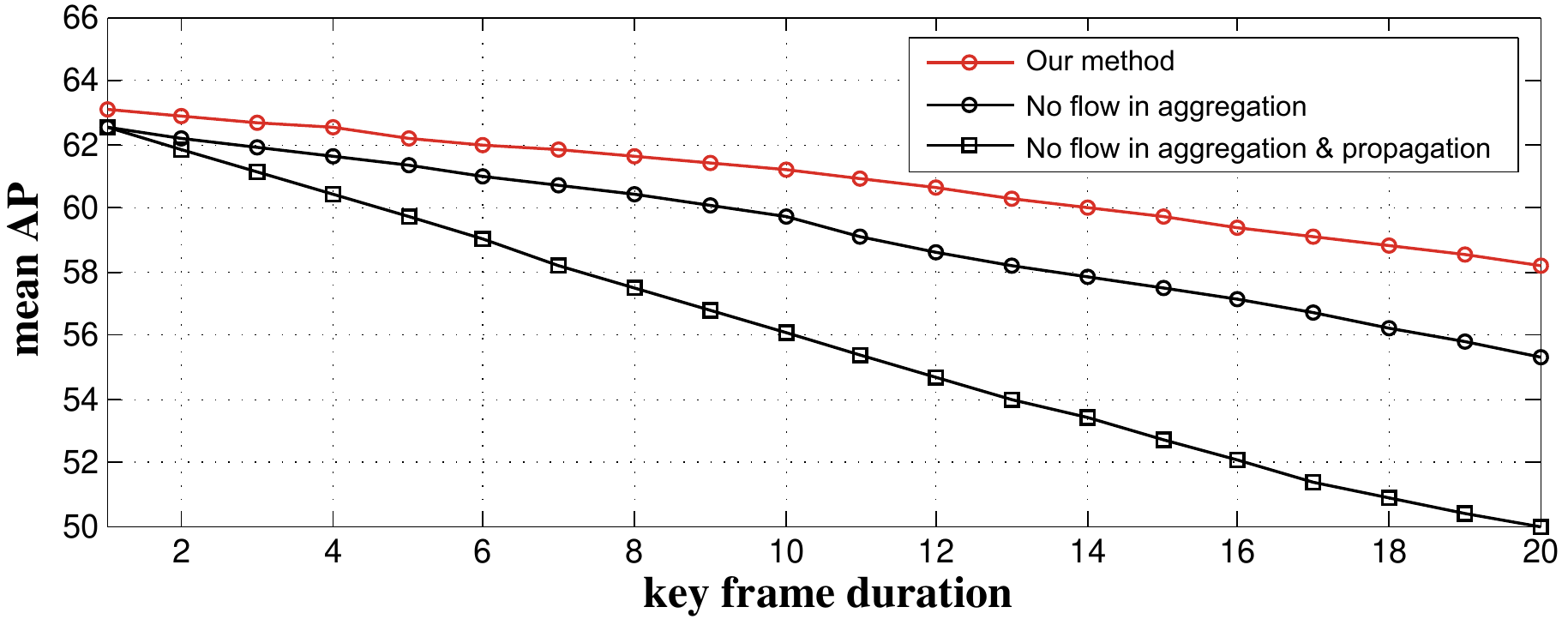}
	\end{center}
	\vspace{-1.5em}
	\caption{Ablation on the effect of flow guidance in flow-guided GRU and in sparse feature propagation. }
	\vspace{-1.5em}
	\label{fig.ablation_aggregation_flow}
\end{figure}

Figure~\ref{fig.ablation_aggregation_flow} shows the speed-accuracy curves of our method with and without flow guidance. The curve is drawn by adjusting the key frame duration $l$. We can see that the curve with flow guidance surpasses that without flow guidance. The performance gap is more obvious when the key frame duration increases (1.5\% mAP score gap at $l = 10$, 2.9\% mAP score gap at $l=20$). This is because the spatial disparity is more obvious when the key frame duration is long. It is worth noting that the accuracy further drops if no flow is applied even for sparse feature propagation on the non-key frames.

Table~\ref{table.ablation_gru} presents the results of training and inference on frame sequences of varying lengths. We tried training on sequences of 2, 4, 8, 16, and 32 frames. The trained network is either applied on trimmed sequences of the same length as in training, or on the untrimmed video sequences without specific length restriction. The experiment suggests that it is beneficial to train on long sequences, but the gain saturates at length 8. Inference on the untrimmed video sequences leads to accuracy on par with that of trimmed, and can be implemented easier. By default, we train on sequences of length 8, and apply the trained network on untrimmed sequences.

Table~\ref{table.ablation_aggregation} further compares the proposed flow-guided GRU method with the feature aggregation approach in~\cite{zhu2017towards}. An mAP score of 58.4\% is achieved by the aggregation approach in~\cite{zhu2017towards}, which is comparable with the single frame baseline at 6.5$\times$ theoretical speedup. But it is still 2.8\% shy in mAP of utilizing flow-guided GRU, at close computational overhead.  

We further studied several design choices in flow-guided GRU. $\phi$ function with ReLU nonlinearity leads to 3.9\% higher mAP score compared to $\tanh$ nonlinearity. The ReLU nonlinearity seems to converge faster than $\tanh$ in our network. If computation allows, it would be more efficient to increase the accuracy by making the flow-guided GRU module wider (1.2\% mAP score increase by enlarging channel width from 128-d to 256-d), other than by stacking multiple layers of the flow-guided GRU module (accuracy drops when stacking 2 or 3 layers).

\setlength{\tabcolsep}{5pt}
\renewcommand{\arraystretch}{1.2}
\begin{table}[t]
	\begin{center}
		\begin{tabular}{l | c | c | c | c |c }
			\hline
			train sequence length & 2 & 4 & 8 & 16 & 32 \\
			\hline
			inference trimmed, mAP (\%) & 59.5 & 61.0 & 61.5 & 61.6 & 61.5 \\
			\hline
			inference untrimmed, mAP (\%) & 56.4 & 60.6 & 61.2 & 61.4 & 61.5 \\
			\hline
		\end{tabular}
	\end{center}
	\caption{Ablation of sequence length in training and inference.}
	\vspace{-1.5em}
	\label{table.ablation_gru}
\end{table}

\setlength{\tabcolsep}{2pt}
\renewcommand{\arraystretch}{1}
\begin{table}[t]
	\begin{center}
		\begin{tabular}{l | c | c | c }
			\hline
			aggregation method & mAP (\%) & params (M) & FLOPs (B) \\
			\hline
			single frame baseline& 58.3 & 5.6 & 2.39 \\
			\hline
			feature aggregation in~\cite{zhu2017towards} & 58.4 & 8.3 & 0.37 \\
			\hline
			\hline
			\textbf{GRU (128-d, default)}& 61.2 & 9.0 & 0.41 \\
			\hline
			GRU (256-d)& 62.4 & 13.0 & 0.64 \\
			\hline
			GRU (tanh for $\phi$)& 57.3 & 9.0 & 0.41 \\
			\hline
			\hline
			GRU (stacking 2 layers)& 61.4 & 9.9 & 0.47 \\
			\hline
			GRU (stacking 3 layers)& 60.6 & 10.8 & 0.53 \\
			\hline
		\end{tabular}
	\end{center}
	\caption{Ablation on feature aggregation.}
	\vspace{-1.5em}
	\label{table.ablation_aggregation}
\end{table}

\subsubsection{Accurate Realtime Video Object Detection on Mobile}

\begin{figure}
	\begin{center}
		\includegraphics[width=.8\linewidth]{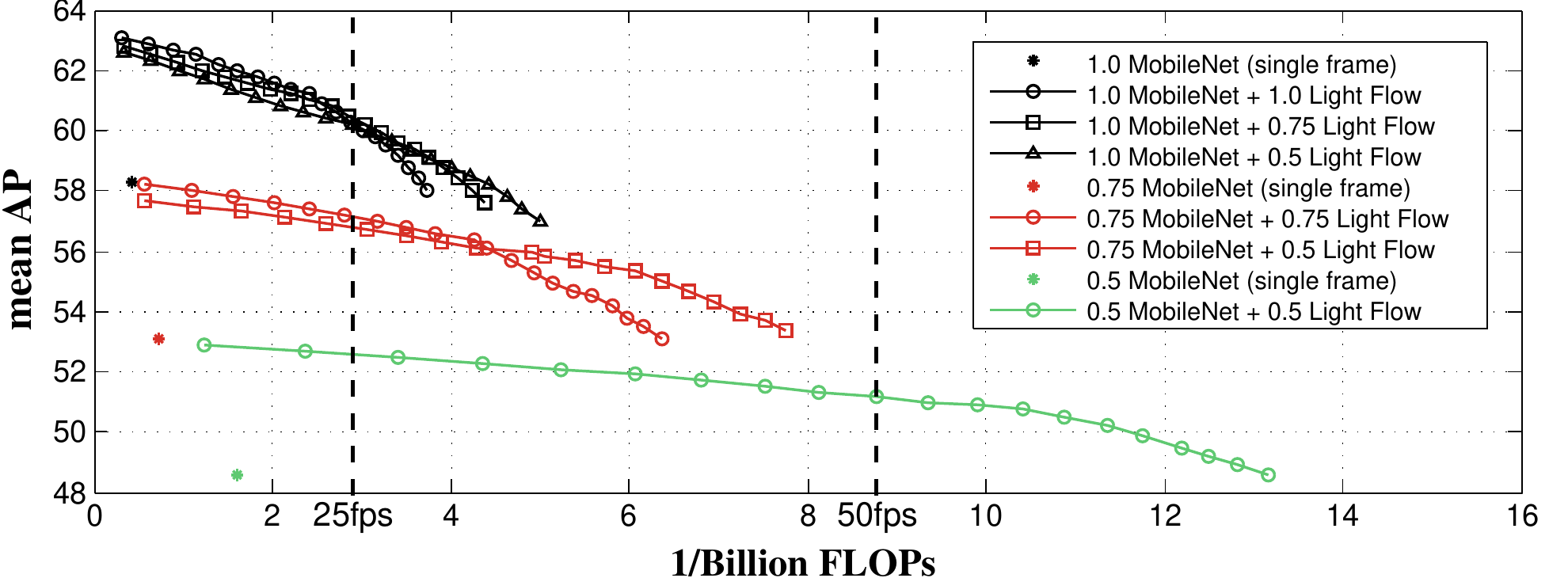}
	\end{center}
	\vspace{-1.5em}
	\caption{Speed-accuracy trade-off curves of our method utilizing networks of different computational complexity. Curves are drawn with different key frame duration length $l \in \{1,2,3,...,20\}$.}
	\label{fig.ablation_our_method_key_frame_duration}
\end{figure}

Figure~\ref{fig.ablation_our_method_key_frame_duration} presents the speed-accuracy trade-off curve of our method, drawn with varying key frame duration length $l$ from 1 to 20. Multiple curves are presented, which correspond to networks of different complexity ($\alpha \times \beta \in \{1.0, 0.75, 0.5\} \times \{1.0, 0.75, 0.5\}$). When $l=1$, the image recognition network is densely applied on each frame, as in the single frame baseline. The difference is flow-guided GRU is applied. The derived accuracy by such dense feature aggregation is noticeably higher than that of the single frame baseline. With increased key frame duration length, the accuracy drops gracefully as the computation overhead relieves. The accuracy of our method at long duration length ($l=20$) is still on par with that of the single frame baseline, and is 10.6$\times$ more computationally efficient. The above observation holds for the curves of networks of different complexity.

As for comparison of different curves, we observe that under adequate computational power, networks of higher complexity ($\alpha=1.0$) would lead to better speed-accuracy tradeoff. On the other hand, networks of lower complexity ($\alpha=0.5$) would perform better under limited computational power.

At our mobile test platform, the proposed system achieves an accuracy of 60.2\% at speed of 25.6 frames per second ($\alpha=1.0$, $\beta=0.5$, $l=10$). The accuracy is 51.2\% at a frame rate of 50Hz ($\alpha=0.5$, $\beta=0.5$, $l=10$). Table~\ref{table.ablation_our_method} summarizes the results. 

\setlength{\tabcolsep}{2pt}
\renewcommand{\arraystretch}{1}
\begin{table}[t]
	\begin{center}
		\begin{tabular}{l | c | c | c | c }
			\hline
			method & mAP (\%) & Params (M) & FLOPs (B) & runtime (fps) \\
			\hline
			Single frame baseline ($\alpha = 1.0$)& 58.3 & 5.6 & 2.39 & 4.0 \\
			\hline
			Single frame baseline ($\alpha = 0.75$)& 53.1 & 3.4 & 1.36 & 7.6 \\
			\hline
			Single frame baseline ($\alpha = 0.5$)& 48.6 & 1.7 & 0.62 & 16.4 \\
			\hline
			\hline
			Our method ($\alpha = 1.0$, $\beta = 1.0$) & 61.2 & 9.0 & 0.41 & 12.5 \\
			\hline
			Our method ($\alpha = 1.0$, $\beta = 0.75$) & 60.8 & 7.8 & 0.37 & 18.2 \\
			\hline
			Our method ($\alpha = 1.0$, $\beta = 0.5$) & 60.2 & 7.1 & 0.34 & 25.6 \\
			\hline
			Our method ($\alpha = 0.75$, $\beta = 0.75$) & 56.4 & 5.3 & 0.23 & 26.3 \\
			\hline
			Our method ($\alpha = 0.75$, $\beta = 0.5$) & 56.0 & 4.6 & 0.20 & 37.0 \\
			\hline
			Our method ($\alpha = 0.5$, $\beta = 0.5$) & 51.2 & 2.6 & 0.11 & 52.6 \\
			\hline
		\end{tabular}
	\end{center}
	\caption{Speed-accuracy performance of our method.}
	\vspace{-1.5em}
	\label{table.ablation_our_method}
\end{table}

\section{In Context of Previous Work on Mobile}

There are also some other endeavors trying to make object detection efficient enough for devices with limited computational power. They can be mainly classified into two major branches: lightweight image object detectors making the per-frame object detector fast, and mobile video object detectors exploiting temporal information.

\subsection{Lightweight Image Object Detector}\label{sec.lightweight_image_detector}

In spite of the work towards more accurate object detection by exploiting deeper and more complex networks, there are also efforts designing lightweight image object detectors for practical applications. Of them, the improvements of YOLO~\cite{redmon2016you}, SSD~\cite{liu2016ssd}, together with the lastest Light-head R-CNN~\cite{li2017light} are of the best speed-accuracy trade-off.

YOLO~\cite{redmon2016you} and SSD~\cite{liu2016ssd} are one-stage object detectors, where the detection result is directly produced by the network in a sliding window fashion. YOLO frames object detection as a regression problem, and a light-weight detection head directly predicts bounding boxes on the whole image. In YOLO and its improvements, like YOLOv2~\cite{redmon2016yolo9000} and Tiny YOLO~\cite{darknet13}, specifically designed feature extraction networks are utilized for computational efficiency. For SSD, the output space of bounding boxes are discretized into a set of anchor boxes, which are classified by a light-weight detection head. In its improvements, like SSDLite~\cite{sandler2018inverted} and Tiny SSD~\cite{wong2018tiny}, more efficient feature extraction networks are also utilized.
	
Light-head R-CNN~\cite{li2017light} is of two-stage, where the object detector is applied on a small set of region proposals. In previous two-stage detectors, either the detection head or its previous layer, is of heavy-weight. In Light-head R-CNN, position-sensitive feature maps~\cite{dai2016rfcn} are exploited to relief the burden. It shows better speed-accuracy performance than the single-stage detectors.

Lightweight image object detector is an indispensable component for our video object detection system. On top of it, our system can further significantly improve the speed-accuracy trade-off curve. Here we choose to integrate Light-head R-CNN into our system, thanks to its outstanding performance. Other lightweight image object detectors should be generally applicable within our system.

\subsection{Mobile Video Object Detector}\label{sec.mobile_video_detector}

Despite the practical importance of video object detection on devices with limited computational power, there is scarce literature. Till very recently, there are two latest works seeking to exploit temporal information for addressing this problem.

In Fast YOLO~\cite{shafiee2017fast}, a modified YOLOv2~\cite{redmon2016yolo9000} detector is applied on sparse key frames, and the detected bounding boxes are directly copied to the the non-key frames, as their detection results. Although sparse key frames are exploited for acceleration, no feature aggregation or flow-guided warping is applied. No end-to-end training for video object detection is performed. Without all these important components, its accuracy cannot compete with ours. But direct comparison is difficult, because the paper does not report any accuracy numbers on any datasets for their method, with no public code.

In~\cite{liu2017mobile}, MobileNet SSDLite~\cite{sandler2018inverted} is applied densely on all the video frames, and multiple Bottleneck-LSTM layers are applied on the derived image feature maps to aggregate information from multiple frames. It cannot speedup upon the single-frame baseline without sparse key frames. Extending it to exploit sparse key frame features would be non-trival. It would involve feature alignment, which is also lacking in~\cite{liu2017mobile}. Its performance also cannot be easily compared with ours. It reports accuracy on a subset of ImageNet VID, where the split is not publicly known. Its code is also not public.

Both two systems cannot compete with the proposed system. They both do not align features across frames. Besides, \cite{shafiee2017fast} does not aggregate features from multiple frames for improving accuracy, while~\cite{liu2017mobile} does not exploit sparse key frames for acceleration. Such design choices are vital towards high performance video object detection.

\subsection{Comparison on ImageNet VID}

Of all the systems discussed in Section~\ref{sec.lightweight_image_detector} and Section~\ref{sec.mobile_video_detector}, SSDLite~\cite{sandler2018inverted}, Tiny YOLO~\cite{darknet13}, and YOLOv2~\cite{redmon2016yolo9000} are the most related systems that can be compared at proper effort. They all seek to improve the speed-accuracy trade-off by optimizing the image object detection network. Although they do not report results on ImageNet VID~\cite{russakovsky2015imagenet}, they all public their code fortunately. We first carefully reproduced their results in paper (on PASCAL VOC~\cite{everingham2010pascal} and COCO~\cite{lin2014microsoft}), and then trained models on ImageNet VID, also by utilizing ImageNet VID and ImageNet DET train sets. The trained models are applied on each video frame for video object detection. By varying the input image frame size (shorter side in \{448, 416, 384, 352, 320, 288, 256, 224\} for SSDLite and Tiny YOLO, and \{320, 288, 256, 224, 192, 160, 128\} for YOLO v2), we can draw their speed-accuracy trade-off curves. The technical report of Fast YOLO~\cite{shafiee2017fast} is also very related. But it neither reports accuracy nor has public code. We cannot compare with it. Note that the comparison is at the detection system level. We do not dive into the details of varying technical designs.   

Figure~\ref{fig.single_image_detectors} presents the the speed-accuracy curves of different systems on ImageNet VID validation. For our system, the curve is drawn also by adjusting the image size\footnote{the input image resolution of the flow network is kept to be half of the resolution of the image recognition network.}(shorter side for image object detection network in \{320, 288, 256, 224, 208, 192, 176, 160\}), for fair comparison. The width multipliers $\alpha$ and $\beta$ are set as 1.0 and 0.5 respectively, and the key frame duration length $l$ is set as 10. Our system surpasses all the existing systems by clear margin. Our method achieves an accuracy of 60.2\% at 25.6 fps. Meanwhile, YOLOv2, SSDLite and Tiny YOLO obtain accuracies of 58.7\%, 57.1\%, and 44.1\% at frame rates of 0.3, 3.8 and 2.2 fps respectively. To the best of our knowledge, for the first time, we achieve realtime video object detection on mobile with reasonably good accuracy.

\section{Discussion}

In this paper, we propose a light weight network for video object detection on mobile devices. We verified that the principals of sparse feature propagation and multi-frame feature aggregation also hold at very limited computational overhead. A very small flow network, Light Flow, is proposed. A flow-guided GRU module is proposed for effective feature aggregation.

A possible issue with the current approach is that there would be short latency in processing online streaming videos. Because the recognition on the key frame is still not fast enough. It would be interesting to study this problem in the future.

\clearpage

\bibliographystyle{splncs}
\bibliography{egbib}
\end{document}